\documentclass[letterpaper, 10 pt, conference]{ieeeconf}  

\IEEEoverridecommandlockouts                              

\title{\LARGE \bf
$T^3$ Planner: A Self-Correcting LLM Framework for Robotic Motion Planning with Temporal Logic
}

\author{Jia Li$^{1}$ and Guoxiang Zhao$^{1, *}$}

\usepackage[colorlinks=true,   
            linkcolor=black,    
            anchorcolor=black, 
            citecolor=green,   
            urlcolor=blue,     
            bookmarks=true,    
            pdfstartview=FitH  
           ]{hyperref}
\usepackage{amsmath,amssymb,amsfonts}
\usepackage[section]{placeins}
\usepackage{algorithm}
\usepackage{algorithmic}
\usepackage{graphicx}
\usepackage{textcomp}
\usepackage{xcolor}
\usepackage{booktabs}
\usepackage{multirow}
\usepackage{comment}
\usepackage{array}
\usepackage[T1]{fontenc}
\usepackage{flushend}
\usepackage{tabularx}
\usepackage{cite}

\makeatletter
\def\@cite#1#2{[#1\if@tempswa, #2\fi]}

\makeatother

\begin{document}

\maketitle
\thispagestyle{empty}
\pagestyle{empty}

\begin{abstract}


Translating natural language instructions into executable motion plans is a fundamental challenge in robotics. Traditional approaches are typically constrained by their reliance on domain-specific expertise to customize planners, and often struggle with spatio-temporal couplings that usually lead to infeasible motions or discrepancies between task planning and motion execution. 
Despite the proficiency of Large Language Models (LLMs) in high-level semantic reasoning, hallucination could result in infeasible motion plans. In this paper, we introduce the $T^3$ Planner, an LLM-enabled robotic motion planning framework that self-corrects it output with formal methods. The framework decomposes spatio-temporal task constraints via three cascaded modules, each of which stimulates an LLM to generate candidate trajectory sequences and examines their feasibility via a Signal Temporal Logic (STL) verifier until one that satisfies complex spatial, temporal, and logical constraints is found.
Experiments across different scenarios show that $T^3$ Planner significantly outperforms the baselines. The required reasoning can be distilled into a lightweight Qwen3-4B model that enables efficient deployment. All supplementary materials are accessible at \url{https://github.com/leeejia/T3\_Planner}.

\end{abstract}

\section{Introduction}

The ability to transform natural language instructions into executable motion actions has long been a key objective in the fields of robotics and human-robot interaction. It holds the promise of enabling the widespread adoption of human-robot collaboration in various domains such as home assistance and industrial assembly \cite{c1,c2}. 
However, natural language instructions often involve common sense reasoning, context comprehension, and inherent ambiguity \cite{c5}, necessitating both syntactic parsing and semantic grounding in real world \cite{c6}. 
Natural language instructed robots are additionally required to consider motional constraints in finding actional plans. 
How to build a robust and generalizable language-to-motion system is inherently a formidable challenge. 

Recent research highlights the capacity of Large Language Models (LLMs) in robotic reasoning and planning \cite{c9,c10}. 
However, independent LLM-based planners suffer from uncertainty issues such as hallucinations \cite{c15}, where physically impossible movements are commonly identified in the outputs of such planners. 
Additionally, limited situational awareness \cite{c16}, such as ignoring obstacles, often leads to suboptimal, infeasible or even unsafe solutions and thus poses great threat to real-world robotic systems in the application of LLM-based planners.
In order to address these issues, verification and feedback mechanisms have been proposed for LLM-enabled planners. Rethinking \cite{c18,c19} generates textual explanations of past failures to prevent recurring issues, but its significant dependencies on historical data accumulation and feedback quality limits its computational scalability and generalization capabilities. 
Regeneration approaches~\cite{c21,c23} synthesize new plans through environmental observation analysis and adapt well to environmental changes but its applicability depends on high-precision perceptions and computationally intensive processing.
Replanning~\cite{c25,c26} uses closed-loop feedback to facilitate real-time failure detection and adjustment.
It effectively deals with dynamic or uncertain environments at the cost of substantial computational overhead and high-feedback quality.

In this work, we introduce $T^3$ Planner, a closed-loop cascaded planning framework based on LLMs and formal verification. The hierarchical framework decomposes the natural language instructions at different granularities and sequentially addresses spatial, temporal, logic and motional constraints specified in the instructions. 
Our method shows superior performance compared to baselines in three different scenarios.
The main contributions of this work are as follows: 1) we propose the $T^3$ Planner, a novel framework that iteratively verifies the outputs of cascaded LLMs with Signal Temporal Logic (STL) and it solves natural language instructed robotic motion planning tasks with multiple constraints. 2) We introduce a task decomposition method that breaks down implicit and coupled spatio-temporal natural language instructed tasks that off-the-shelf LLMs can solve within finite time. 3) We also provide a lightweight deployment solution via distilling the task reasoning capabilities required by the framework into the Qwen3-4B model.

\section{Related Works}

\subsection{Task and motion planning}
In recent years, various formal language models have been proposed to encode planning problems, such as Linear Temporal Logic (LTL) and STL. 
AutoTAMP \cite{c33} and Cook2LTL \cite{Cook2LTL} translate natural language into STL and LTL respectively, to generate unambiguous, temporally precise robot plans.
Compared to natural language, these symbolic formal languages exhibit significant advantages in knowledge density and structural regularity \cite{c27}, effectively eliminating the inherent ambiguity of natural language.

\subsection{LLMs for task and motion planning} 

LLM-enabled task planners break down tasks into point-to-point navigation subproblems, solved using existing tools like A*, RRT*, or LQR \cite{lavalle2006planning}, forming a sequential switching system. LLM-GROP \cite{LLM-GROP} uses LLMs to define symbolic goals and object placements for semantic rearrangement, processed by a classical TAMP planner. LLM+P \cite{LLM+P} reformulates planning problems within the PDDL framework for classical task planning solutions. These methods offer low offline computation and solving complexity but often produce infeasible subtask sequences for nonlinear systems and require replanning due to disturbances, leading to significant online computational overhead.

LLM-enabled motion planners utilize LLMs to search for the mapping from state or observation to control and can be divided into three types: LLM-as-controller, LLM-as-path-planner and LLM-as-controller-generator.
LLM-as-controller method directly employs LLMs as the mapping from observations to control signals \cite{RT-2, SayTap}. While this method effectively processes semantic environmental information, it suffers from poor real-time performance due to extensive computations and network latency. 
LLM-as-path-planner method primarily leverage LLMs to produce intermediate representations of control commands based on natural language instructions. \cite{GPIR2023, song2023llm} plan a sequence of subgoals, grounded on specific robotic states such as configurations and velocities, followed by a low-level controller that enables the robot to sequentially track the subgoals.
LLM-as-controller-generator method uses LLMs as a controller generator and outputs the controller script that can be executed online as a conventionally designed controller \cite{chen2024roboscript, guo2024castl}.
The latter two methods offer high online efficiency but heavily rely on the reasoning capabilities of LLM models and supporting modules like perception and control, limiting its effectiveness to prestructured environments.

\subsection{Natural language instruction to STL specification}
In recent years, various studies have explored the automatic translation of natural language into STL using LLMs. NL2TL \cite{chen-etal-2023-nl2tl} combines GPT-3-generated data with T5 model fine-tuning. 
KGST \cite{fang-etal-2025-enhancing} introduces external knowledge to refine rough translations, significantly reducing the misinterpretation rate of complex nested temporal formulas. 
Inspired by these approaches, this paper adopts a more lightweight strategy of few-shot LLM end-to-end generation with human verification, where Gemini-2.5-pro is prompted directly to translate natural language descriptions into STL, and syntax and semantics verifications are imposed by annotators familiar with formal methods.

\section{Problem Formulation}

Consider a nonlinear robot in the two-dimensional space, where the robot's objective is to interpret natural language instructions that can be further articulated as STL formulas and translate them into feasible robotic motion trajectories that satisfy multiple constraints.

Throughout this paper, we consider instructions that invlove a) spatial constraints, such as the regions where the robot must or must not visit; b) temporal constraints, where the robot shall finish certain tasks by deadlines or within given durations; c) logical constraints, where the robot shall finish subtasks following certain sequential orderings, conditional branching, or implicit prerequisites.

The instructions can be translated into STL formulas.
An STL formula is recursively composed of atomic predicates, logical operators ($\lnot$, $\land$, $\lor$, $\rightarrow$, $\leftrightarrow$), and temporal operators ($F$, $G$, $U$). Given a signal $s(t)$ and an STL formula $\varphi$, its quantitative semantics $\rho(s,t,\varphi)$ measures the degree to which the signal satisfies the formula at time $t$, often referred to as the robustness score \cite{robustness09}. In this paper, we define atomic predicates such as $\text{in}(A)$ to indicate that the robot is within region $A$. 



\section{Method}

\begin{figure*}[h]%
\vspace{1.4mm}
\centerline{\includegraphics[width=0.95\linewidth]{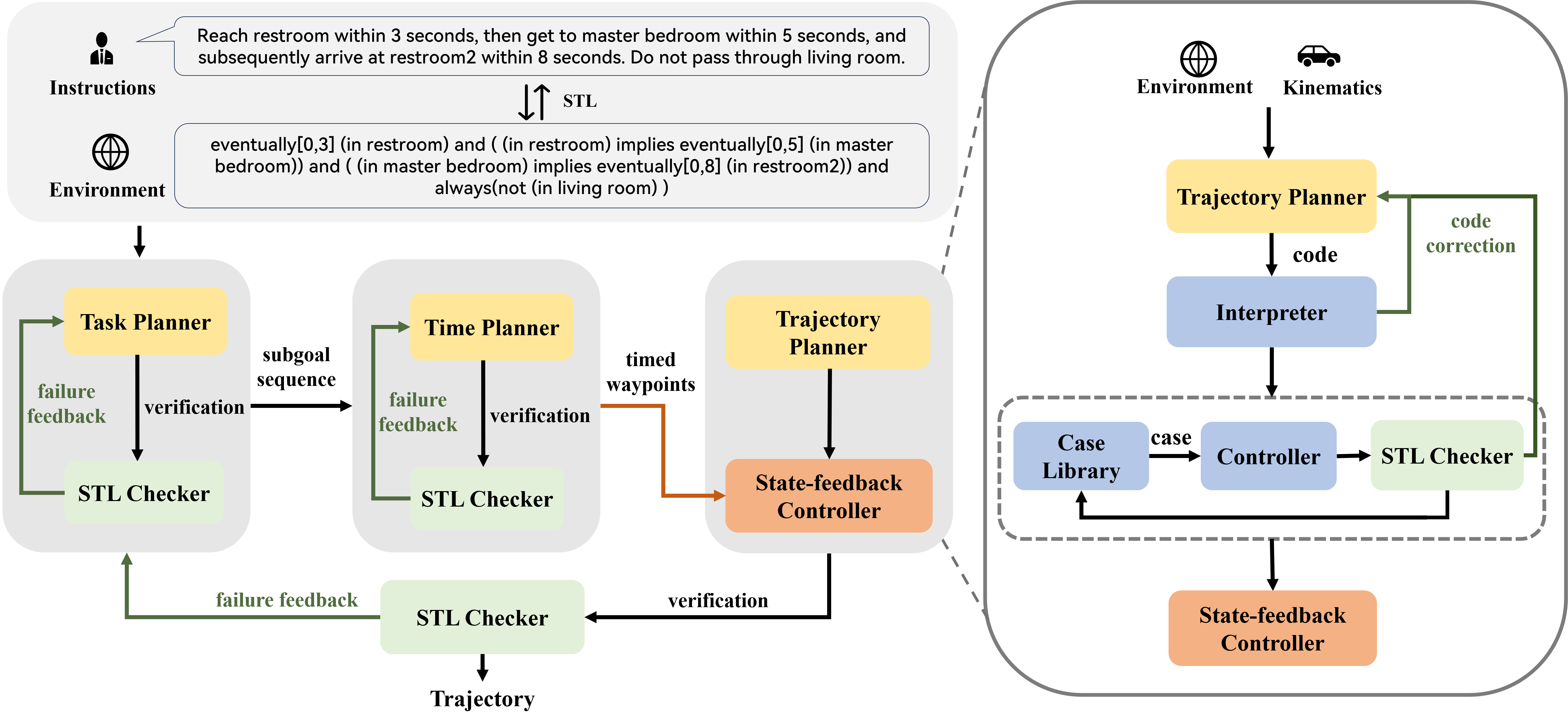}}
\vspace{-2ex}
\caption{Illustration of $T^3$ Planner}
\label{fig:framework}
\end{figure*}

$T^3$ Planner is a cascaded architecture with each level fed back with validations to enhance the feasibility and robustness of the planned trajectory. The framework diagram of $T^3$ Planner is shown in Fig. \ref{fig:framework}.

\subsection{Overall description}
The framework translates natural language instructions with three cascaded modules, each addressing distinct aspects of task execution. The \textbf{Task Planner} interprets spatial constraints from the instructions and generate a sequence of subgoal waypoints, ensuring considerations such as obstacle avoidance and target region selection. Following this, the \textbf{Time Planner} assigns each subgoal with a timestamp, adhering to temporal constraints such as overall task duration and allowed operating temporal window of each subtask. The \textbf{Trajectory Planner} then crafts a motion controller to formulate kinematically feasible trajectories from these timed waypoints. However, LLMs suffer from hallucinations, and the results of a single planning may be erroneous. Therefore, it is imperative to verify the output of LLMs. We propose a cyclic verification mechanism that corresponds to these three planning modules. These loops are crucial for ensuring compliance and maintaining the robustness of the overall system: \textbf{Logical Verification} checks the extent to which spatial constraints and logical consistency are satisfied after task planning. In the event of any violations, it will trigger the regeneration of subgoal sequence; \textbf{Temporal Verification} evaluates the spatio-temporal constraints of timed waypoints. When violations occur, it will lead to replanning of the task; and \textbf{Motional Verification} conducts a feasibility check on the final trajectory. If the verification fails, the entire process will be restarted from the task planner.

\subsection{Planning with LLMs}

We prompt LLM to generate motion controller in text format. We adopt Chain-of-Thought (CoT) and few-shot prompting to stimulate the latent capabilities of LLMs.

\begin{figure}[htbp]
\centerline{\includegraphics[width=\linewidth]{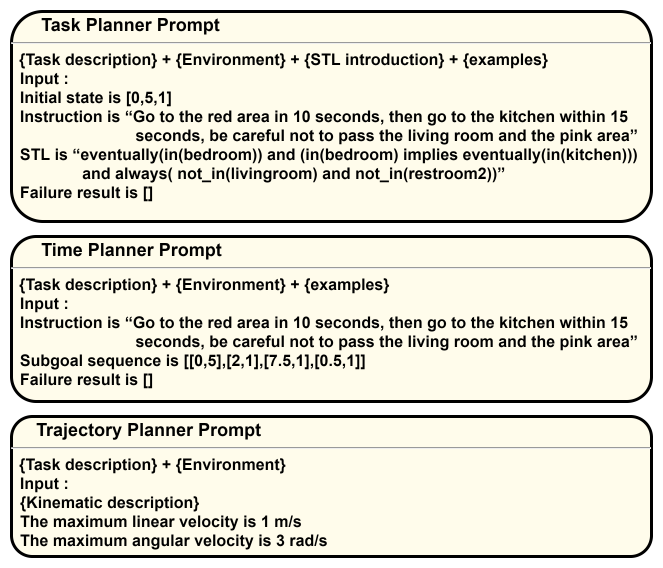}}
\vspace{-2ex}\caption{Prompt templetes of $T^3$ Planner}
\label{fig:prompt}
\end{figure}

As illustrated in Fig. \ref{fig:prompt}, the prompt inputs for the task and time planner include: (i) a description of the task and the scenario environment; (ii) the input and output formats; (iii) few-shot examples. The task planner prompt also includes an introduction to the STL specification. In addition, when the initial planning fails and enters the iterative generation loop, the associated output will be fed back to LLM, prompting the LLM to reflect on the causes of the errors and regenerate better result. The prompt sent to the trajectory planner include: (i) a description of the task; (ii) a description of the kinematic model; (iii) the output format. The trajectory planner generates a controller, which is then simulated to produce the final trajectory.
Each of these three modules can be represented by the following formula: 
\begin{equation}\begin{aligned}
    subgoal\_seq &= \text{Task\_Plan}(I, \varphi, S_0, \psi) ,\end{aligned}\end{equation}
    \begin{equation}\begin{aligned}
    timed\_waypoints &= \text{Time\_Plan}(I, subgoal\_seq, \psi) ,\end{aligned}\end{equation}
    \begin{equation}\begin{aligned}
    controller &= \text{Trajectory\_Plan}(\chi ,v_{\max}, \omega_{\max}),
\end{aligned}\end{equation}
where $I$ denotes the natural language instruction, $S_0$ denotes the initial state, $\varphi$ denotes the STL formula, and $\psi$ denotes the previous failed planning result, initialized as a null value. The kinematic model description is summarized in $\chi$, and the controller's maximum allowable linear and angular velocities are denoted by $v_{\max}$ and $\omega_{\max}$ respectively.

\subsection{Verifications to guide LLMs}

\textbf{Form of Verification Mechanism:}
The robot's objective requires its continuous operations to satisfy both spatial and temporal constraints. To verify compliance with these requirements, we employ STL for formal verification thanks to its native support for both spatial and temporal constraints over real-valued signals. The STL's quantitative semantics allow us to not only check satisfaction of safety constraints but also measure their robustness margin. This is particularly valuable for time-critical operations where the timing of spatial transitions directly impacts mission success.

We analyze sampled measurements of the robot's position over time; i.e., a discrete sequence of timed waypoints $\tau=((x_1,y_1,t_1),(x_2,y_2,t_2),\cdots,(x_n,y_n,t_n))$, 
where each tuple $(x_i,y_i)\in\mathbb{R}^2$ denotes a spatial coordinate and $t_i\in\mathbb{R}_{\geq0}$ denotes the timestamp when the position is sampled.
The asymptotic consistency of discrete time evaluation is established in \cite{robustness09}, where sufficiently large sampling frequency contributes to negligible evaluation errors. 



\textbf{Logical and Temporal Verification:}
The outputs of different modules could be infeasible. To address this, verification loops are introduced to mitigate error propagation.

The subgoal sequence generated by the \textbf{Task Planner} may incur traversing obstacle areas, incorrect target positioning, and wrong regional visit sequences. To verify its correctness, \textbf{Logical Verification} employs the STL robustness, evaluated with respect to the associated STL specification ignoring temporal constraints, to check the spatial compliance of the subgoal sequence. If the robustness score $\rho < 0$, it indicates an unsatisfied task planning, and the \textbf{Task Planner} will regenerate a new subgoal sequence based on the previous failed outcomes. 

The timed waypoints produced by the \textbf{Time Planner} may suffer from irrational time planning that does not comply with the STL specification.
\textbf{Temporal Verification} assesses the temporal feasibility of the timed waypoints through the vanilla STL checking, where $\rho < 0$ triggers the \textbf{Time Planner} to replan the time allocation.

\textbf{Motional Verification:}
The output of the \textbf{Trajectory Planner} module is a python script functioned as the motion controller, and its successful execution is subject to python syntax, logical correctness and functional alignment, and a single iteration of controller generation is not surely complied with all these requirements. Towards this end, a dual-layer \textbf{Motional Verification} is employed for iterative optimization, as illustrated in the magnified region of Fig. \ref{fig:framework}.

The first layer involves static syntax verification, where the generated controller code undergoes pre-compilation analysis using a python code interpreter. If syntax errors such as missing symbols or structural anomalies are detected, \textbf{Trajectory Planner} is triggered to regenerate a controller.

The second layer is functional verification that examines the satisfiability of the simulated trajectories given the generated controller under a set of typical cases. 
The controller first generates motion trajectories using timed waypoints from a case library as input parameters and then performs STL robustness verification with a returned robustness score $\rho\geq0$ serving as the passing criterion. If a trajectory violates constraints ($\rho < 0$), the system triggers \textbf{Trajectory Planner} to regenerate the controller. This process iterates until the generated controller passes all verification cases. 
The case library is composed of expert-annotated timed waypoints paired with STL specifications. 
It covers a wide range of scenarios by adjusting parameters such as time constraints, obstacle positions and speed limits, and it also focuses on edge cases such as tight simulation deadlines and complex obstacle layouts. 
This facilitates complete coverage and constraint validity of the case set. 

\section{Experiments}

\subsection{Task scenarios}
\begin{figure*}[htbp]
\centerline{\includegraphics[width=1\linewidth]{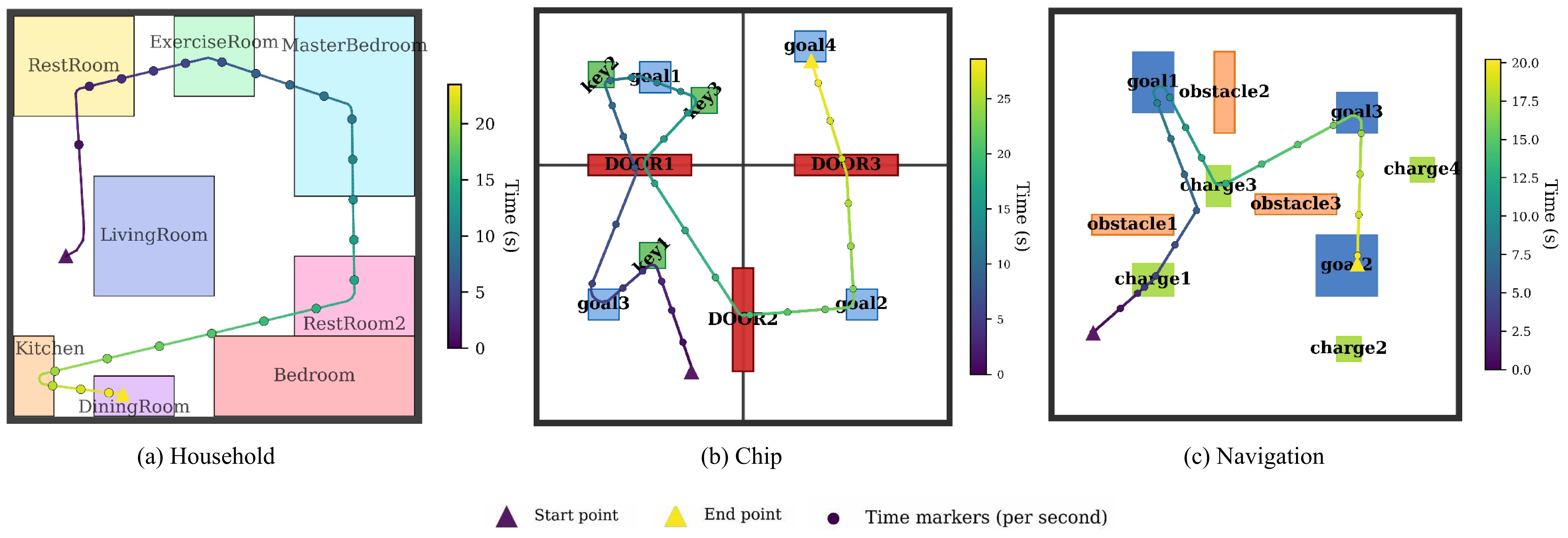}}
\vspace{-2ex}\caption{ $T^3$ Planner produces feasible motion trajectories. (a) In the household scenario, the instruction is ``Visit each room once, except for the Bedroom and Living Room, and complete the whole process within 30 seconds.'' (b) In the chip scenario, the instruction is ``Strive to reach all the goals. At the same time, always avoid touching the walls. Complete the process in less than 30 seconds.'' (c) In the navigation scenario, the instruction is ``Seek to reach every single goal. Also, always make sure not to touch the obstacles. Complete the process in less than 30 seconds.''}
\label{fig:scenario}
\end{figure*}

We evaluate $T^3$ Planner in three distinct task scenarios, each configured in a simulated 2D environment where the primary objective is robotic navigation. 
The scenario cases are shown in Fig. \ref{fig:scenario}. 

\subsubsection{Household}
A household environment \cite{c43}, consists eight rooms with labeled spatial regions. The robot must navigate through or avoid specific rooms based on natural language instructions containing spatio-temporal constraints.

\subsubsection{Chip}
This is a sequential task scenario adapted from Chip's Challenge, consists of multiple rooms, and the robot can enter each room only when it has collected the right key to the room. The robot should not touch the walls and should follow the instructions to reach one or more goals.

\subsubsection{Navigation}
A battery-powered robot navigates in a cluttered environment to reach its destination and charging station. The robot must always avoid all obstacles and must recharge at a charging station for 3 seconds before pursuing any goals. 
The robot will dynamically select the nearest charging station based on its current position. The robot shall visit specified destinations.

\subsection{Experimental Setups}

\subsubsection{Dataset and Task Complexity}

\textbf{\\ Data Generation: } The distillation training and test datasets for all scenarios were constructed by prompting Gemini-2.5-Pro. Each training instance consists of a natural language instruction, the corresponding reasoning chain, and the final result. Each test instance pairs a natural language instruction with its associated STL specification.
All generated samples underwent manual verification to ensure quality and correctness. The size of the dataset is detailed in Table \uppercase\expandafter{\romannumeral1}.

\begin{table}[H]
\centering
\small
\caption{Number of Dataset Samples in three scenarios}
\begin{tabularx}{\linewidth}{lXXX}
\toprule
\textbf{} & \textbf{Household} & \textbf{Chip} & \textbf{Navigation} \\
\midrule
Train Dataset & 2\,451 & 929 & 976 \\
Test Dataset     &   400  & 300 & 300 \\
\bottomrule
\end{tabularx}
\end{table}

\noindent
\textbf{Instruction Classification: } To systematically evaluate $T^3$ planner's capability across task difficulty levels, we partition the test set instructions into three categories: simple, intermediate, and advanced. The classification is based on the number of constraints, the complexity of temporal relationships, and the logical depth of the instruction. This allows for a more fine-grained analysis of model performance.
\begin{itemize}
\item Simple instructions come with a straightforward structure that contain one or two constraint types; without conditional logic, all constraints are explicit and mutually independent.

\item Intermediate instructions consist of moderate complexity that typically involve two or three constraint types; may contain explicit conditionals or implicit constraints, must be satisfied simultaneously.  

\item Advanced instructions combine multiple constraint types and typically include conditionals, implicit constraints, or a combination of both, demanding multi-step reasoning and tightly constrained execution.

\end{itemize}

\subsubsection{Baselines} 

We select AutoTAMP \cite{c33} as the baseline method. In AutoTAMP, a pre-trained LLM serves as a translator, converting natural language instructions into STL specifications. These specifications are subsequently compiled by an STL motion planner to synthesize timestamped waypoints $(x_i, y_i, t_i)$.
Different from AutoTAMP's translation-oriented approach, our framework emphasizes closed-loop planning that combines multi-layer planning with STL robustness verification and failure feedback.

\subsubsection{Proposed Method Variants}
Our framework's performance is evaluated using several powerful LLMs as its core reasoning engine, including GPT-4o, DeepSeek-reasoner, and Gemini-2.5-Pro. We also test a lightweight, distilled version to assess its deployment feasibility.

\subsubsection{Implementation and Distillation Details}
\textbf{ \\STL Tools: }The calculation of the STL robustness score adopts the PSY-TaLiRo toolbox \cite{psy-taliro}. This python toolkit provides a complete set of interfaces for parsing the semantics of spatio-temporal logic. 

\noindent
\textbf{Knowledge Distillation: }We adopted a teacher-student knowledge distillation approach to enhance the reasoning skills of a compact model. We designated Gemini-2.5-Pro as the teacher and Qwen3-4B as the student.

The teacher model was first prompted to generate detailed reasoning chains and final answers for each sample in our training set. This created a high-quality dataset for supervised fine-tuning. Then we fine-tuned the Qwen3-4B student model on this generated data using the LLaMA-Factory. The training runs for 5 epochs with a learning rate of 2e-5. The experiments were conducted on a computer with 4 × NVIDIA RTX 4090 (24GB) GPUs, utilizing NCCL for distributed training. 

\subsection{Results} 
We evaluate the planner's feasibility and computational efficiency on the test sets presented in Table \uppercase\expandafter{\romannumeral1}, which comprise 400, 300, and 300 samples from the household, chip, and navigation scenarios, respectively.
Performance indicators include: (i) Success Rate (SR): ratio of tasks completed with a trajectory satisfying all spatio-temporal and kinematic constraints; (ii) Average Call Time (AT): the average time in  seconds per call to LLM; (iii) Average Calls (AC): the average number of LLM invocations per instruction needed to obtain a valid plan.

\textbf{Main Performance Evaluation: } As shown in Table \uppercase\expandafter{\romannumeral2}, our framework achieves robust task success rates that outperforms AutoTAMP, achieving an improvement in success rate of 30–40\%.  
When employing reasoning-oriented models such as DeepSeek-reasoner and Gemini-2.5-Pro, the framework attains a success rate exceeding 97\% in chip scenarios and surpasses 93\% in navigation scenarios. Even with the lightweight distilled  Qwen3-4B model, the success rate remains above 92\%.

In contrast, the baseline AutoTAMP, which employs LLM as a translator to convert natural language instructions into STL specifications prior to motion planning, exhibited significantly lower performance. 
Our replication experiments utilizing GPT-4o reveal that the majority of failures originate from translation errors in converting natural language to STL specifications, demonstrating inherent limitations of LLMs in translating complex logical constraints.
Our experimental analysis reveals that AutoTAMP is plagued by the issue of suboptimal decision-making, which leads to inefficient solutions. Specifically, AutoTAMP retrieves the specific key for a certain door only when that door becomes the next immediate obstacle in chip scenarios. This inefficiency poses a significant risk of violating the temporal constraints specified in the task, and it may result in the failure of execution. 
\vspace{-0.2cm}
\begin{table}[ht]
\vspace{1.3mm}
\centering
\caption{task success rate of three scenarios}
\label{table:SR}
\begin{tabular}{lccc}
\toprule
 & \textbf{Household} & \textbf{Chip} & \textbf{Navigation} \\
\midrule
AutoTAMP(gpt-4o)&$56.67\%$ & $61.29\%$& $50.00\%$ \\

Ours(gpt-4o)& $91.65\%$ & $91.60\%$ & $83.69\%$ \\

Ours(DeepSeek-reasoner)& $\textbf{95.12\%}$& $97.71\%$& $93.68\%$ \\

Ours(Gemini-2.5-Pro)& $94.74\%$& $\textbf{98.12\%}$& $\textbf{97.03\%}$ \\

Ours(distilled Qwen3-4B)& $ 92.52 \%$& $ 97.17 \%$& $ 92.26   \%$ \\
\bottomrule
\end{tabular}

\smallskip
\footnotesize 
\end{table}

\begin{figure}[htbp]
\centerline{\includegraphics[width=1.0\linewidth]{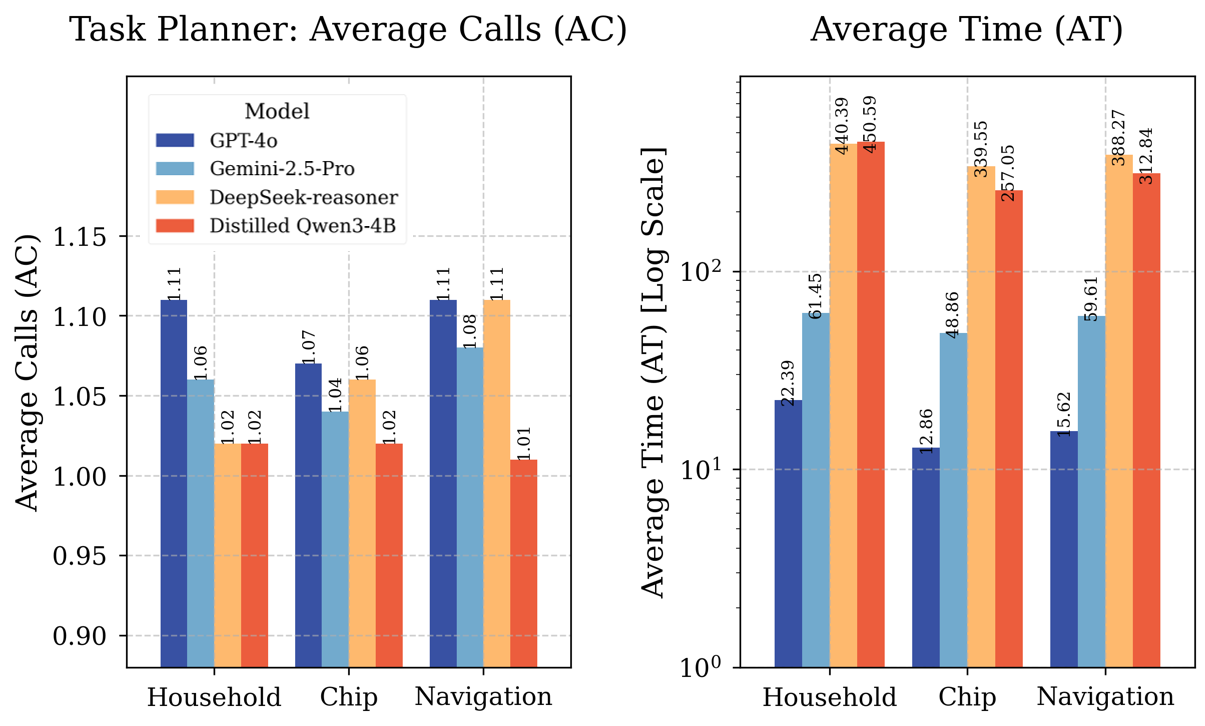}}
\vspace{-2ex}\caption{Computational efficiency of task planner}
\label{fig:task_planner}
\end{figure}

\begin{figure}[htbp]
\centerline{\includegraphics[width=1.0\linewidth]{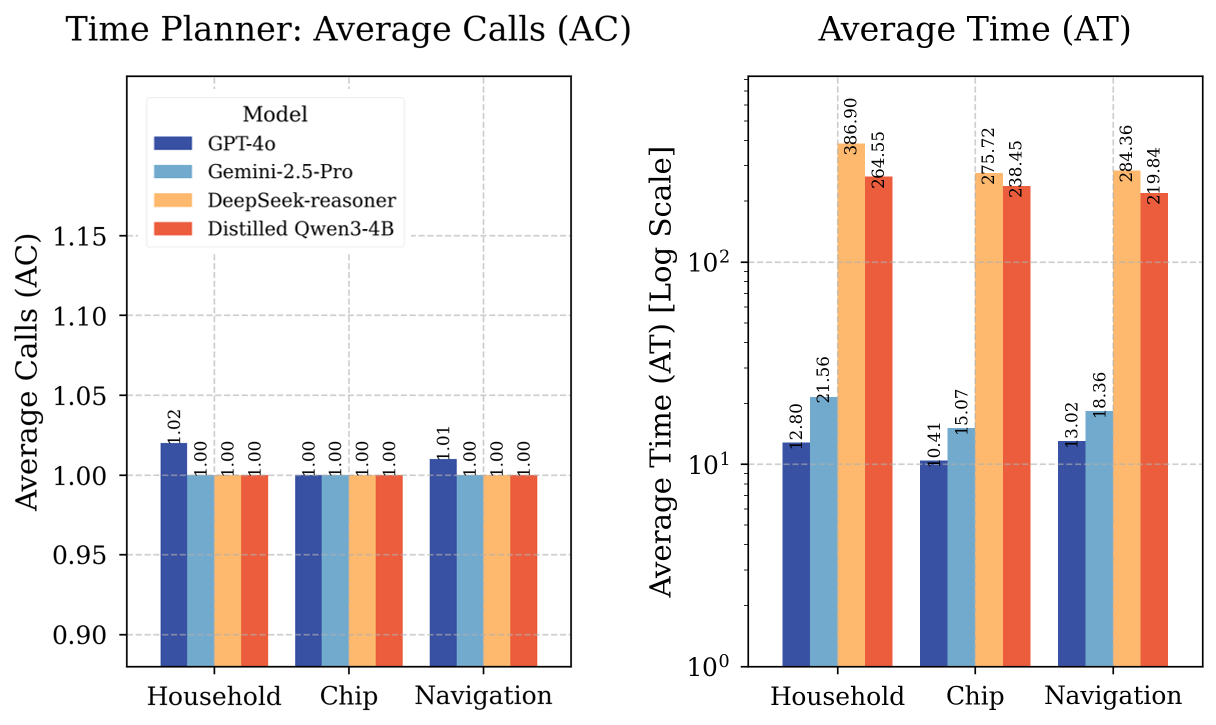}}
\vspace{-2ex}\caption{Computational efficiency of time planner}
\label{fig:time_planner}
\end{figure}

\textbf{Computational Efficiency Analysis: }
We further analyzed the computational efficiency of different models in both task planner and time planner shown in 
Fig. \ref{fig:task_planner} and \ref{fig:time_planner}.
All models attain high single-shot planning success rates, with the distilled Qwen3-4B exhibiting the best empirical performance.
In addition, both GPT-4o and Gemini-2.5-Pro achieve a favorable accuracy–efficiency balance, whereas a single call to DeepSeek-reasoner consumes approximately twenty times the latency of GPT-4o.
This show that quantization and acceleration techniques can be used to retune the distillation hyper-parameters and reduce runtime despite of its relative higher per-call latency.

\textbf{Trajectory Planner Evaluation: }
Table \uppercase\expandafter{\romannumeral3} summarizes the performance of the trajectory planner in generating valid controllers. 
Based on 10 independent trials, this efficiency is attributed to the code generation capability of LLMs and the case library-driven verification mechanism that iteratively checks candidate controllers against timed waypoints and STL, reducing redundant LLM calls and thus improving the generation efficiency.
\vspace{-0.1cm}
\begin{table}[h]
\centering
\caption{Performance evaluation of Trajectory Planner}
\label{table:results}
\begin{tabular}{lcccc}
\toprule
 & \textbf{SR} & \textbf{AC} & \textbf{AT} \\
\midrule
Trajectory planner     & 100.00\%     & 2.20  & 18.49  \\
\bottomrule
\end{tabular}
\smallskip
\footnotesize 
\end{table}

\vspace{-4ex}
\begin{table}[H]
\centering
\scriptsize
\setlength{\tabcolsep}{3pt}
\caption{Model performance on atomic and composite tasks}
\label{tab:model_comparison}
\begin{tabular}{@{}lccp{3.6cm}@{}}
\toprule
Model & Atomic & Composite & Failure-mode summary\\[-1pt]
      & SR   & SR      & \\
\midrule
GPT-4o           & 100.0\% & 60.0\% & Fails on multi-constraint and TSP-style optimization.\\
DeepSeek-reasoner      & 100.0\% & 80.0\% & Tends to overlook implicit constraints.\\
Gemini-2.5-Pro   & 100.0\% & 85.0\% & Occasionally misses implicit constraints in complex tasks.\\
distilled Qwen3-4B         & 100.0\% & 80.0\% & Occasionally misses implicit constraints and underperforms on TSP-style optimization.\\
\bottomrule
\end{tabular}
\end{table}

\textbf{LLM Model Capability Evaluation: }
Grounding the task to the household scenario, we create a lightweight test dataset aiming at identifying the capability boundary of each model. The set contains 10 atomic instructions (simple/intermediate) and 20 composite instructions (intermediate/advanced).
Table \uppercase\expandafter{\romannumeral4} summarizes the performance of different LLMs. All models performed reliably on atomic tasks with an accuracy of $100\%$.
However, performance of different models bifurcates on composite-logic tasks, where non-reasoning models like GPT-4o score only $60\%$ compared to other reasoning-oriented ones' $\geq80\%$. 
Notably, our lightweight distilled model obtains an accuracy closely to those full-tank reasoning models.

\begin{figure}[htbp]
\centerline{\includegraphics[width=1.0\linewidth]{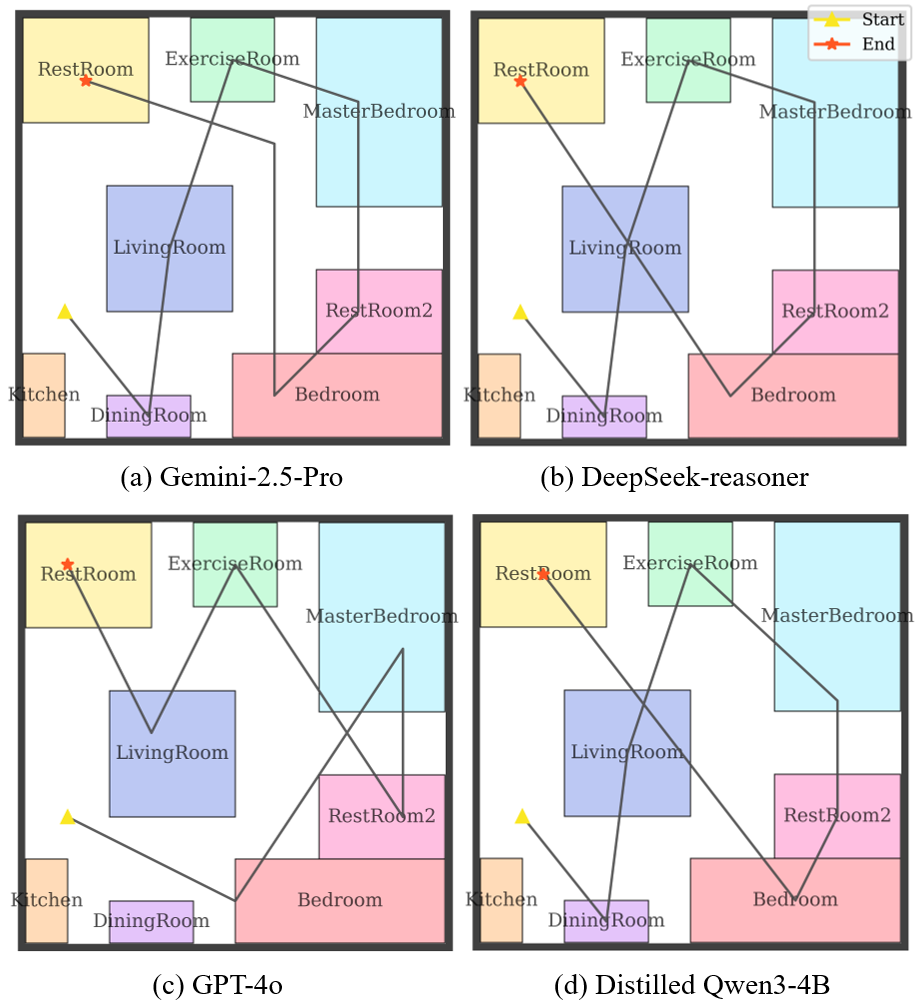}}
\vspace{-2ex}\caption{Trajectories generated by $T^3$ Planner with different LLMs. The instruction is as follows: ``The starting point is the DiningRoom. Within 9 seconds, you must be inside either the LivingRoom or the Bedroom. Should you enter the LivingRoom, you are required to reach the ExerciseRoom within 6 seconds. Should you enter the Bedroom, you must get to the MasterBedroom within 6 seconds. After completing this stage, visit all other rooms, but never enter the Kitchen. Your final destination must be the RestRoom. Complete this entire route in less than 38 seconds, visiting each room a maximum of one time.''}
\label{fig:scenario_2}
\end{figure}

Fig. \ref{fig:scenario_2} illustrates the trajectories generated by $T^3$ Planner using different models before kinematic modeling in the household scenario.
Gemini-2.5-Pro enables the robot to exactly follow the instruction. DeepSeek-reasoner and distilled Qwen3-4B satisfy most constraints but violate an implicit requirement ``visit each room a maximum of one time'' by re-entering the LivingRoom on the path to the RestRoom. GPT-4o fails to satisfy these constraints altogether.

\vspace{-0.5cm}

\subsection{Ablation study} 

\begin{figure}[h]
\centerline{\includegraphics[width=0.8\linewidth]{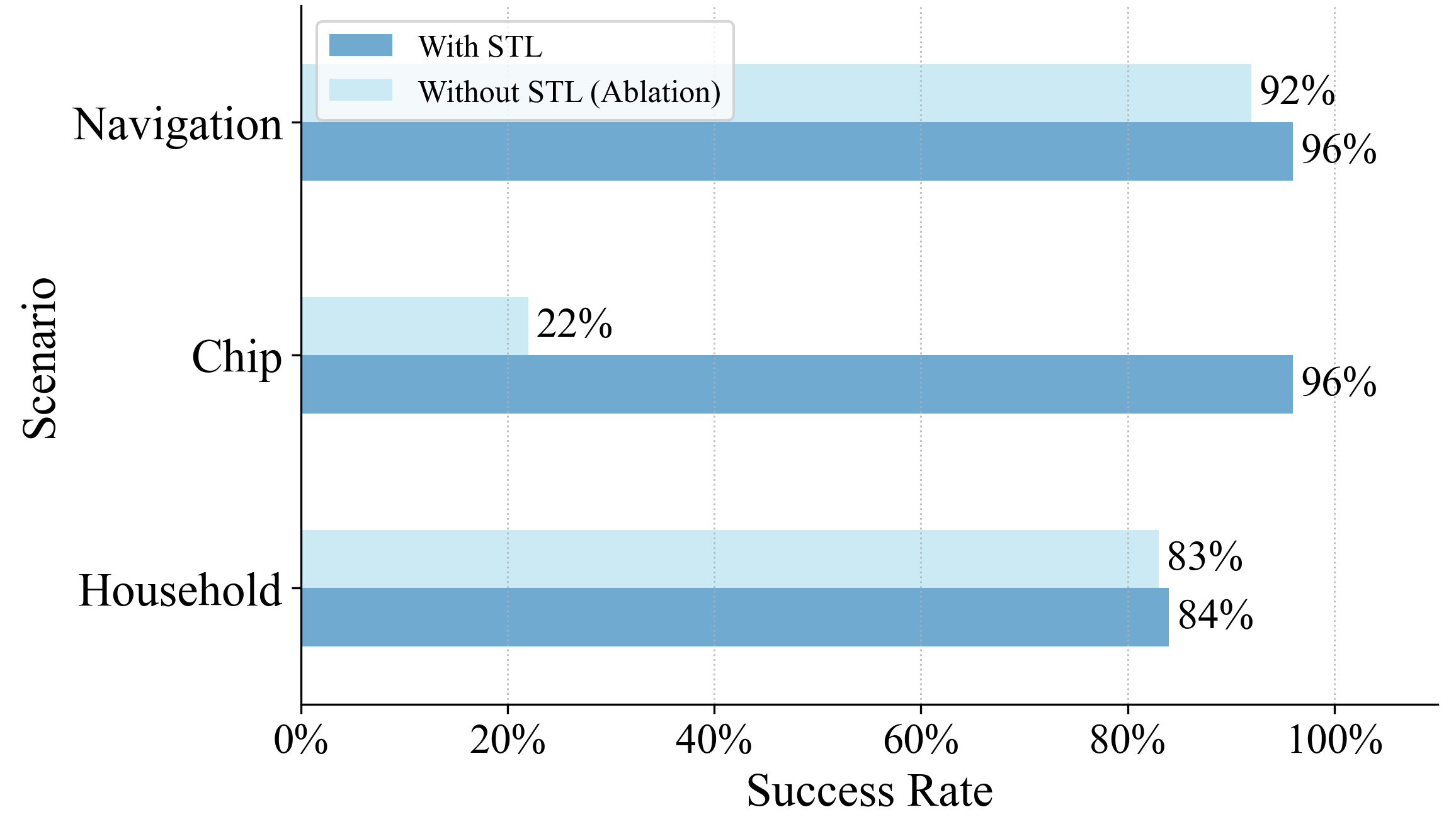}}
\vspace{-0.3cm}
\caption{STL ablation study}
\label{fig:ablation}
\end{figure}

We conduct an ablation study using the GPT-4o task planner to justify the necessity of STL specifications in $T^3$ Planner. We remove the STL specifications from the prompts in three scenarios, each tested on distinct 100 instances. Fig.~\ref{fig:ablation} shows that removing STL specifications contributes to lower success rates across all scenarios. The impact is significant in the chip scenario, where the success rate plummets from $96\%$ to $22\%$, while in the other two scenarios it declines only marginally.
This difference highlights the value of formal specifications for sequentially constrained tasks, where the symbolic structure of STL specifications bounds the LLM search space and the STL input curbs the instructional ambiguity of the model.

\vspace{-2ex}
\vspace{-2ex}
\section{Conclusion}
This paper presents $T^3$ Planner, a feedback framework that integrates LLM high-level reasoning with STL formal verification for robotic motion planning.
Evaluations in three scenarios show that $T^3$ Planner outperforms baseline in feasibility and constraint satisfaction, while the distilled Qwen3-4B variant retains comparable accuracy.
Future directions include extending the framework to multi-robot collaboration and incorporating formal verification as learnable parameter weights into the model.

\bibliographystyle{IEEEtranBST/IEEEtranS}
\bibliography{IEEEtranBST/ref}

\begin{thebibliography}{10}
\providecommand{\url}[1]{#1}
\csname url@rmstyle\endcsname
\providecommand{\newblock}{\relax}
\providecommand{\bibinfo}[2]{#2}
\providecommand\BIBentrySTDinterwordspacing{\spaceskip=0pt\relax}
\providecommand\BIBentryALTinterwordstretchfactor{4}
\providecommand\BIBentryALTinterwordspacing{\spaceskip=\fontdimen2\font plus
\BIBentryALTinterwordstretchfactor\fontdimen3\font minus \fontdimen4\font\relax}
\providecommand\BIBforeignlanguage[2]{{%
\expandafter\ifx\csname l@#1\endcsname\relax
\typeout{** WARNING: IEEEtran.bst: No hyphenation pattern has been}%
\typeout{** loaded for the language `#1'. Using the pattern for}%
\typeout{** the default language instead.}%
\else
\language=\csname l@#1\endcsname
\fi
#2}}

\bibitem{c1}
K.~Lin, C.~Agia, T.~Migimatsu, M.~Pavone, and J.~Bohg, ``Text2motion: From natural language instructions to feasible plans,'' \emph{Autonomous Robots}, vol.~47, no.~8, pp. 1345--1365, 2023.

\bibitem{c2}
S.~Huang, Z.~Jiang, H.~Dong, Y.~Qiao, P.~Gao, and H.~Li, ``Instruct2act: Mapping multi-modality instructions to robotic actions with large language model,'' \emph{arXiv preprint arXiv:2305.11176}, 2023.

\bibitem{c5}
B.~Cai, X.~Ding, Z.~Sun, B.~Qin, T.~Liu, L.~Shang, \emph{et~al.}, ``Self-supervised logic induction for explainable fuzzy temporal commonsense reasoning,'' in \emph{Proceedings of the AAAI Conference on Artificial Intelligence}, vol.~37, no.~11, 2023, pp. 12\,580--12\,588.

\bibitem{c6}
Y.~Liu, W.~Chen, Y.~Bai, X.~Liang, G.~Li, W.~Gao, and L.~Lin, ``Aligning cyber space with physical world: A comprehensive survey on embodied ai,'' \emph{arXiv preprint arXiv:2407.06886}, 2024.

\bibitem{c9}
W.~X. Zhao, K.~Zhou, J.~Li, T.~Tang, X.~Wang, Y.~Hou, Y.~Min, B.~Zhang, J.~Zhang, Z.~Dong, \emph{et~al.}, ``A survey of large language models,'' \emph{arXiv preprint arXiv:2303.18223}, vol.~1, no.~2, 2023.

\bibitem{c10}
T.~Kojima, S.~S. Gu, M.~Reid, Y.~Matsuo, and Y.~Iwasawa, ``Large language models are zero-shot reasoners,'' \emph{Advances in neural information processing systems}, vol.~35, pp. 22\,199--22\,213, 2022.

\bibitem{c15}
J.~Kaddour, J.~Harris, M.~Mozes, H.~Bradley, R.~Raileanu, and R.~McHardy, ``Challenges and applications of large language models,'' \emph{arXiv preprint arXiv:2307.10169}, 2023.

\bibitem{c16}
S.~Kambhampati, ``Can large language models reason and plan?'' \emph{Annals of the New York Academy of Sciences}, vol. 1534, no.~1, pp. 15--18, 2024.

\bibitem{c18}
Z.~Liu, A.~Bahety, and S.~Song, ``Reflect: Summarizing robot experiences for failure explanation and correction,'' in \emph{Proceedings of The 7th Conference on Robot Learning}, ser. Proceedings of Machine Learning Research, vol. 229.\hskip 1em plus 0.5em minus 0.4em\relax PMLR, 06--09 Nov 2023, pp. 3468--3484.

\bibitem{c19}
S.~S. Raman, V.~Cohen, I.~Idrees, E.~Rosen, R.~Mooney, S.~Tellex, and D.~Paulius, ``Cape: Corrective actions from precondition errors using large language models,'' in \emph{2024 IEEE International Conference on Robotics and Automation (ICRA)}, 2024, pp. 14\,070--14\,077.

\bibitem{c21}
Y.~Chen, J.~Arkin, Y.~Zhang, N.~Roy, and C.~Fan, ``Scalable multi-robot collaboration with large language models: Centralized or decentralized systems?'' in \emph{2024 IEEE International Conference on Robotics and Automation (ICRA)}.\hskip 1em plus 0.5em minus 0.4em\relax IEEE, 2024, pp. 4311--4317.

\bibitem{c23}
X.~Zhao, M.~Li, C.~Weber, M.~B. Hafez, and S.~Wermter, ``Chat with the environment: Interactive multimodal perception using large language models,'' in \emph{2023 IEEE/RSJ International Conference on Intelligent Robots and Systems (IROS)}.\hskip 1em plus 0.5em minus 0.4em\relax IEEE, 2023, pp. 3590--3596.

\bibitem{c25}
W.~Huang, F.~Xia, T.~Xiao, H.~Chan, J.~Liang, P.~Florence, A.~Zeng, J.~Tompson, I.~Mordatch, Y.~Chebotar, \emph{et~al.}, ``Inner monologue: Embodied reasoning through planning with language models,'' in \emph{Conference on Robot Learning}.\hskip 1em plus 0.5em minus 0.4em\relax PMLR, 2023, pp. 1769--1782.

\bibitem{c26}
F.~Joublin, A.~Ceravola, P.~Smirnov, F.~Ocker, J.~Deigmoeller, A.~Belardinelli, C.~Wang, S.~Hasler, D.~Tanneberg, and M.~Gienger, ``Copal: corrective planning of robot actions with large language models,'' in \emph{2024 IEEE International Conference on Robotics and Automation (ICRA)}.\hskip 1em plus 0.5em minus 0.4em\relax IEEE, 2024, pp. 8664--8670.

\bibitem{c33}
Y.~Chen, J.~Arkin, C.~Dawson, Y.~Zhang, N.~Roy, and C.~Fan, ``Autotamp: Autoregressive task and motion planning with llms as translators and checkers,'' in \emph{2024 IEEE International conference on robotics and automation (ICRA)}.\hskip 1em plus 0.5em minus 0.4em\relax IEEE, 2024, pp. 6695--6702.

\bibitem{Cook2LTL}
A.~Mavrogiannis, C.~Mavrogiannis, and Y.~Aloimonos, ``Cook2ltl: Translating cooking recipes to ltl formulae using large language models,'' \emph{2024 IEEE International Conference on Robotics and Automation (ICRA)}, pp. 17\,679--17\,686, 2023.

\bibitem{c27}
F.~Xu, Z.~Wu, Q.~Sun, S.~Ren, F.~Yuan, S.~Yuan, Q.~Lin, Y.~Qiao, and J.~Liu, ``Symbol-llm: Towards foundational symbol-centric interface for large language models,'' in \emph{Proceedings of the 62nd Annual Meeting of the Association for Computational Linguistics (Volume 1: Long Papers)}, 2024, pp. 13\,091--13\,116.

\bibitem{lavalle2006planning}
S.~M. LaValle, \emph{Planning algorithms}.\hskip 1em plus 0.5em minus 0.4em\relax Cambridge university press, 2006.

\bibitem{LLM-GROP}
Y.~Ding, X.~Zhang, C.~Paxton, and S.~Zhang, ``Task and motion planning with large language models for object rearrangement,'' \emph{2023 IEEE/RSJ International Conference on Intelligent Robots and Systems (IROS)}, pp. 2086--2092, 2023.

\bibitem{LLM+P}
B.~Liu, Y.~Jiang, X.~Zhang, Q.~Liu, S.~Zhang, J.~Biswas, and P.~Stone, ``Llm+p: Empowering large language models with optimal planning proficiency,'' \emph{ArXiv}, vol. 2304.11477, 2023.

\bibitem{RT-2}
B.~Zitkovich, T.~Yu, S.~Xu, P.~Xu, T.~Xiao, F.~Xia, J.~Wu, P.~Wohlhart, S.~Welker, A.~Wahid, \emph{et~al.}, ``Rt-2: Vision-language-action models transfer web knowledge to robotic control,'' in \emph{Conference on Robot Learning}.\hskip 1em plus 0.5em minus 0.4em\relax PMLR, 2023, pp. 2165--2183.

\bibitem{SayTap}
Y.~Tang, W.~Yu, J.~Tan, H.~Zen, A.~Faust, and T.~Harada, ``Saytap: Language to quadrupedal locomotion,'' in \emph{Proceedings of The 7th Conference on Robot Learning}, ser. Proceedings of Machine Learning Research, vol. 229.\hskip 1em plus 0.5em minus 0.4em\relax PMLR, 06--09 Nov 2023, pp. 3556--3570.

\bibitem{GPIR2023}
Y.~Du, O.~Watkins, Z.~Wang, C.~Colas, T.~Darrell, P.~Abbeel, A.~Gupta, and J.~Andreas, ``Guiding pretraining in reinforcement learning with large language models,'' in \emph{International Conference on Machine Learning}.\hskip 1em plus 0.5em minus 0.4em\relax PMLR, 2023, pp. 8657--8677.

\bibitem{song2023llm}
C.~H. Song, J.~Wu, C.~Washington, B.~M. Sadler, W.-L. Chao, and Y.~Su, ``Llm-planner: Few-shot grounded planning for embodied agents with large language models,'' in \emph{Proceedings of the IEEE/CVF international conference on computer vision}, 2023, pp. 2998--3009.

\bibitem{chen2024roboscript}
J.~Chen, Y.~Mu, Q.~Yu, T.~Wei, S.~Wu, Z.~Yuan, Z.~Liang, C.~Yang, K.~Zhang, W.~Shao, \emph{et~al.}, ``Roboscript: Code generation for free-form manipulation tasks across real and simulation,'' \emph{arXiv preprint arXiv:2402.14623}, 2024.

\bibitem{guo2024castl}
W.~Guo, Z.~Kingston, and L.~E. Kavraki, ``Castl: Constraints as specifications through llm translation for long-horizon task and motion planning,'' \emph{arXiv preprint arXiv:2410.22225}, 2024.

\bibitem{chen-etal-2023-nl2tl}
Y.~Chen, R.~Gandhi, Y.~Zhang, and C.~Fan, ``{NL}2{TL}: Transforming natural languages to temporal logics using large language models,'' in \emph{Proceedings of the 2023 Conference on Empirical Methods in Natural Language Processing}.\hskip 1em plus 0.5em minus 0.4em\relax Association for Computational Linguistics, Dec. 2023, pp. 15\,880--15\,903.

\bibitem{fang-etal-2025-enhancing}
Y.~Fang, Z.~Jin, J.~An, H.~Chen, X.~Chen, and N.~Zhan, ``Enhancing transformation from natural language to signal temporal logic using {LLM}s with diverse external knowledge,'' in \emph{Findings of the Association for Computational Linguistics: ACL 2025}.\hskip 1em plus 0.5em minus 0.4em\relax Vienna, Austria: Association for Computational Linguistics, July 2025, pp. 10\,446--10\,458.

\bibitem{robustness09}
G.~E. Fainekos and G.~J. Pappas, ``Robustness of temporal logic specifications for continuous-time signals,'' \emph{Theoretical Computer Science}, vol. 410, no.~42, pp. 4262--4291, 2009.

\bibitem{c43}
C.~Finucane, G.~Jing, and H.~Kress-Gazit, ``Ltlmop: Experimenting with language, temporal logic and robot control,'' in \emph{2010 IEEE/RSJ International Conference on Intelligent Robots and Systems}.\hskip 1em plus 0.5em minus 0.4em\relax IEEE, 2010, pp. 1988--1993.

\bibitem{psy-taliro}
Q.~Thibeault, J.~Anderson, A.~Chandratre, and G.~Pedrielli, ``Psy-taliro: A python toolbox for search-based test generation for cyber-physical systems,'' in \emph{Formal Methods for Industrial Critical Systems}.\hskip 1em plus 0.5em minus 0.4em\relax Springer International Publishing, 2021, pp. 223--231.

\end{thebibliography}


\begin{thebibliography}{99}

\bibitem{c1} K. Lin, C. Agia, T. Migimatsu, M. Pavone, J. Bohg, Text2Motion: from natural language instructions to feasible plans, Auton. Robots. 47 (2023) 1345–1365.
 

\bibitem{c2} S. Huang, Z. Jiang, H. Dong, Y. Qiao, P. Gao, H. Li, Instruct2Act: Mapping Multi-modality Instructions to Robotic Arm Actions with Large Language Model, 2024.----It has been accepted by ICLR2024, but how can I find the cite of the conference

\bibitem{c3} Hu H, Chen J, Liu H et al (2022) Natural language-based automatic programming for industrial robots. J Grid Comput 20:26.

\bibitem{c4} Nyga D, Roy S, Paul R et al (2018) Grounding robot plans from natural language instructions with incomplete world knowledge. In: Proceedings of the conference on robot learning (CoRL), pp 714–723.

\bibitem{c5} B. Cai, X. Ding, Z. Sun, B. Qin, T. Liu, B. Wang, L. Shang, Self-Supervised Logic Induction for Explainable Fuzzy Temporal Commonsense Reasoning, in: AAAI Conference on Artificial Intelligence, 2023.

\bibitem{c6} Y. Liu, W. Chen, Y. Bai, J.-H. Luo, X. Song, K. Jiang, Z. Li, G. Zhao, J. Lin, G. Li, W. Gao, L. Lin, Aligning Cyber Space with Physical World: A Comprehensive Survey on Embodied AI, ArXiv. abs/2407.06886 (2024).

\bibitem{c7} C. R. Garrett, R. Chitnis, R. Holladay, B. Kim, T. Silver, L. P. Kaelbling, and T. Lozano-Pe´ rez. “Integrated task and motion planning”. In: Annual review of control, robotics, and autonomous systems 4.1 (2021), pp. 265–293.

\bibitem{c8} S. Li, D. Park, Y. Sung, et al., "Reactive task and motion planning under temporal logic specifications," in Proceedings of the IEEE international conference on robotics and automation (ICRA), 2021, pp. 12618–12624. 


\bibitem{c9} W. X. Zhao, K. Zhou, J. Li, T. Tang, X. Wang, Y. Hou, Y. Min, B. Zhang, J. Zhang, Z. Dong, et al., “A survey of large language models,” arXiv preprint arXiv:2303.18223, 2023.

\bibitem{c10} T. Kojima, S. S. Gu, M. Reid, Y. Matsuo, and Y. Iwasawa, “Large language models are zero-shot reasoners,” Advances in Neural Information Processing Systems (NeurIPS), vol. 35, pp. 22199–22213, 2022.

\bibitem{c11} W. Huang, P. Abbeel, D. Pathak, and I. Mordatch, “Language models as zero-shot planners: Extracting actionable knowledge for embodied agents,” in International Conference on Machine Learning (ICML), 2022.

\bibitem{c12} Q. Dong, L. Li, D. Dai, C. Zheng, Z. Wu, B. Chang, X. Sun, J. Xu, and Z. Sui, “A survey for in-context learning,” arXiv preprint arXiv:2301.00234, 2022.

\bibitem{c13} S. Li, X. Puig, C. Paxton, Y. Du, C. Wang, L. Fan, T. Chen, D.A. Huang, E. Akyurek,¨ A. Anandkumar, et al., “Pre-trained language models for interactive decision-making,” Advances in Neural Information Processing Systems (NeurIPS), vol. 35, pp. 31199–31212, 2022.

\bibitem{c14} E. Lavrinovics, R. Biswas, J. Bjerva, K. Hose, Knowledge Graphs, Large Language Models, and Hallucinations: An NLP Perspective, 2024.

\bibitem{c15} Kaddour J, Harris J, Mozes M et al (2023) Challenges and applications of large language models. pp 1–72. arXiv preprint arXiv:2307.10169.

\bibitem{c16} S. Kambhampati. “Can large language models reason and plan?”In: Annals of the New York Academy of Sciences 1534.1 (2024), pp. 15–18.


\bibitem{c17} Lin BY, Fu Y, Yang Ketal (2023a) Swiftsage: a generative agent with fast and slowthinking for complexinteractive tasks. In: Con ference on neural information processing systems (NeurIPS), pp 23813–23825.

\bibitem{c18} Liu Z, Bahety A, Song S (2023f) Reflect: Summarizing robot experiences for failure explanation and correction. In: Proceed ings of the conference on robot learning (CoRL), pp 3468–3484.

\bibitem{c19} Raman SS, Cohen V, Paulius D et al (2023) CAPE: corrective actions from precondition errors using large language models. In: CoRL workshop on language and robot learning: language as grounding, pp 1–9.

\bibitem{c20} Wang Z, Cai S, Liu A et al (2023d) Describe, explain, plan and select: interactive planning with llms enables open-world multi task agents. In: Conference on neural information processing systems (NeurIPS), pp 34153–34189.

\bibitem{c21} Chen Y, Arkin J, Zhang Y et al (2024c) Scalable multi-robot collaboration with large language models: centralized or decen tralized systems? In: Proceedings of the IEEE international conference on robotics and automation (ICRA).

\bibitem{c22} S. Vemprala, R. Bonatti, A. Bucker, et al., "Chatgpt for robotics: design principles and model abilities," arXiv preprint arXiv:2306.17582, 2023, pp. 1–25. 

\bibitem{c23} X. Zhao, M. Li, C. Weber, et al., "Chat with the environment: interactive multimodal perception using large language models," in Proceedings of the IEEE/RSJ international conference on intelligent robots and systems (IROS), 2023, pp. 3590–3596.

\bibitem{c24} Z. Zhao, W. S. Lee, D. Hsu, "Large language models as commonsense knowledge for large-scale task planning," in Conference on neural information processing systems (NeurIPS), 2023, pp. 31967–31987. 

\bibitem{c25} W. Huang, F. Xia, T. Xiao, H. Chan, J. Liang, P. Florence, A. Zeng, J. Tompson, I. Mordatch, Y. Chebotar, et al., “Inner monologue: Embodied reasoning through planning with language models,” in Conference on Robot Learning (CoRL), 2023.

\bibitem{c26} F. Joublin, A. Ceravola, P. Smirnov, F. Ocker, J. Deigmoeller, A. Belardinelli, C. Wang, S. Hasler, D. Tanneberg, M. Gienger, “CoPAL: Corrective Planning of Robot Actions with Large Language Models,” in 2024 IEEE International Conference on Robotics and Automation (ICRA), 2023, pp. 8664–8670. 


\bibitem{c27} F. Xu, Z. Wu, Q. Sun, S. Ren, F. Yuan, S. Yuan, Q. Lin, Y. Qiao, J. Liu, Symbol-LLM: Towards Foundational Symbol-centric Interface For Large Language Models, in: L.-W. Ku, A. Martins, V. Srikumar (Eds.), Proceedings of the 62nd Annual Meeting of the Association for Computational Linguistics (Volume 1: Long Papers), Association for Computational Linguistics, Bangkok, Thailand, 2024: pp. 13091–13116.

\bibitem{c28} Z. Zhao, S. Cheng, Y. Ding, Z. Zhou, S. Zhang, D. Xu, Y. Zhao, A Survey of Optimization-Based Task and Motion Planning: From Classical to Learning Approaches, IEEE/ASME Transactions on Mechatronics. (2024) 1–27.


\bibitem{c29} M.A. Sormoli, K. Koufos, M. Dianati, R. Woodman, A Survey on Hybrid Motion Planning Methods for Automated Driving Systems, ArXiv. abs/2406.05575 (2024).

\bibitem{c30} K. Rana, J. Haviland, S. Garg, J. Abou-Chakra, I. Reid, and N. Suenderhauf, “SayPlan: Grounding Large Language Models using 3D Scene Graphs for Scalable Robot Task Planning,” in Proceedings of The 7th Conference on Robot Learning, Nov. 2023, vol. 229, pp. 23–72. [Online]. Available: https://proceedings.mlr.press/v229/rana23a.html




\bibitem{c31}A. Jiao, T. P. Patel, S. Khurana, et al., "Swarm-gpt: combining large language models with safe motion planning for robot choreography design," in NeurIPS robot learning workshop: pre-training, fine-tuning, and generalization with large scale models, 2023, pp. 1–10. 

\bibitem{c32} W. Huang, C. Wang, R. Zhang, et al., "Voxposer: composable 3D value maps for robotic manipulation with language models," in Proceedings of the conference on robot learning (CoRL), 2023, pp. 540–562. 

\bibitem{c33}Y. Chen, J. Arkin, Y. Zhang, N. A. Roy, C. Fan, “AutoTAMP: Autoregressive Task and Motion Planning with LLMs as Translators and Checkers,” in 2024 IEEE International Conference on Robotics and Automation (ICRA), 2023, pp. 6695–6702. 

\bibitem{c34}S. Wang, M. Han, Z. Jiao, Z. Zhang, Y. Wu, S. -C. Zhu, H. Liu, “LLM3: Large Language Model-based Task and Motion Planning with Motion Failure Reasoning,” in 2024 IEEE/RSJ International Conference on Intelligent Robots and Systems (IROS), 2024, pp. 12086–12092. 


\bibitem{c35}J. Liang, W. Huang, F. Xia, P. Xu, K. Hausman, B. Ichter, P. Florence, A. Zeng, “Code as Policies: Language Model Programs for Embodied Control,” in 2023 IEEE International Conference on Robotics and Automation (ICRA), 2023, pp. 9493–9500. 

\bibitem{c36} I. Singh, V. Blukis, A. Mousavian, A. Goyal, D. Xu, J. Tremblay, D. Fox, J. Thomason, A. Garg, “ProgPrompt: Generating Situated Robot Task Plans using Large Language Models,” in 2023 IEEE International Conference on Robotics and Automation (ICRA), pp. 11523–11530, 2023. 

\bibitem{c37} L. Guan, K. Valmeekam, S. Sreedharan, et al., "Leveraging pre-trained large language models to construct and utilize world models for model-based task planning," in Conference on neural information processing systems (NeurIPS), 2023, pp. 79081–79094. 

\bibitem{c38}J. Wei, X. Wang, D. Schuurmans, M. Bosma, F. Xia, E. Chi, Q. V.
Le, D. Zhou, et al., “Chain-of-thought prompting elicits reasoning in large language models,” Advances in Neural Information Processing Systems (NeurIPS), vol. 35, pp. 24824–24837, 2022.

\bibitem{c39}W. Wu, S. Mao, Y. Zhang, Y. Xia, L. Dong, L. Cui, F. Wei, “Mind's Eye of LLMs: Visualization-of-Thought Elicits Spatial Reasoning in Large Language Models,” in Neural Information Processing Systems, 2024. 

\bibitem{c40} Y. Weng, M. Zhu, F. Xia, B. Li, S. He, S. Liu, B. Sun, K. Liu, J. Zhao, “Large Language Models are Better Reasoners with Self-Verification,” in Findings of the Association for Computational Linguistics: EMNLP 2023, Singapore, 2023, pp. 2550–2575, Association for Computational Linguistics. 

\bibitem{c41} T. B. Brown, B. Mann, N. Ryder, M. Subbiah, J. D. Kaplan, P. Dhariwal, A. Neelakantan, P. Shyam, G. Sastry, A. Askell, S. Agarwal, A. Herbert-Voss, G. Krueger, T. Henighan, R. Child, A. Ramesh, D. M. Ziegler, J. Wu, C. Winter, C. Hesse, M. Chen, E. Sigler, M. Litwin, S. Gray, B. Chess, J. Clark, C. Berner, S. McCandlish, A. Radford, I. Sutskever, D. Amodei, “Language Models are Few-Shot Learners,” in Advances in Neural Information Processing Systems, H. Larochelle, M. Ranzato, R. Hadsell, M. F. Balcan, H. Lin, Eds. Curran Associates, Inc., 2020, vol. 33, pp. 1877–1901.

\bibitem{c42} A. Donze´ and O. Maler, “Robust satisfaction of temporal logic over real-valued signals,” in Formal Modeling and Analysis of Timed Systems: 8th International Conference, FORMATS 2010, Klosterneuburg, Austria, September 8-10, 2010. Proceedings 8, pp. 92–106, Springer, 2010.

\bibitem{c43} C. Finucane, G. Jing, and H. Kress-Gazit, “Ltlmop: Experimenting with language, temporal logic and robot control,” in 2010 IEEE/RSJ International Conference on Intelligent Robots and Systems. IEEE, 2010, pp. 1988–1993.

\end{thebibliography}

\end{document}